\documentclass[letterpaper, 10 pt, conference]{ieeeconf}

\IEEEoverridecommandlockouts

\usepackage{booktabs}
\usepackage{multirow}
\usepackage{graphicx}
\usepackage{microtype}
\usepackage{amsmath,amssymb,amsfonts}
\usepackage{xcolor}
\usepackage{tikz}
\usepackage{textcomp}
\usepackage{lipsum}

\DeclareMathOperator*{\argmin}{arg\,min}
\newcommand{\bb}[1]{{\textbf{#1}}}

\newcommand\copyrighttext{%
  \footnotesize \textcopyright 2023 IEEE. Personal use of this material is permitted.
  Permission from IEEE must be obtained for all other uses, in any current or future
  media, including reprinting/republishing this material for advertising or promotional
  purposes, creating new collective works, for resale or redistribution to servers or
  lists, or reuse of any copyrighted component of this work in other works.}
\newcommand\copyrightnotice{%
\begin{tikzpicture}[remember picture,overlay]
\node[anchor=south,yshift=10pt] at (current page.south) {\fbox{\parbox{\dimexpr\textwidth-\fboxsep-\fboxrule\relax}{\copyrighttext}}};
\end{tikzpicture}%
}

\graphicspath{figures}

\title{\LARGE \bf
Collision-aware In-hand 6D Object Pose Estimation\\using Multiple Vision-based Tactile Sensors
}

\author{Gabriele M. Caddeo$^{1, 2}$, Nicola A. Piga$^{1}$,  Fabrizio Bottarel$^{1}$ and Lorenzo Natale$^{1}$%
\thanks{$^{1}$Istituto Italiano di Tecnologia, Via San Quirico, 19 D, Genova, Italy. {\tt\small name.surname@iit.it}}%
\thanks{$^{2}$Universit\`a di Genova, Via All'Opera Pia, 13, Genova.}%
}

\usepackage[hidelinks]{hyperref}
\usepackage{cite}
\begin{document}

\maketitle
\copyrightnotice
\thispagestyle{empty}
\pagestyle{empty}

\begin{abstract}
In this paper, we address the problem of estimating the in-hand 6D pose of an object in contact with multiple vision-based tactile sensors. We reason on the possible spatial configurations of the sensors along the object surface. Specifically, we filter contact hypotheses using geometric reasoning and a Convolutional Neural Network (CNN), trained on simulated object-agnostic images, to promote those that better comply with the actual tactile images from the sensors. We use the selected sensors configurations to optimize over the space of 6D poses using a Gradient Descent-based approach.
We finally rank the obtained poses by penalizing those that are in collision with the sensors. We carry out experiments in simulation using the DIGIT vision-based sensor with several objects, from the standard YCB model set. The results demonstrate that our approach estimates object poses that are compatible with actual object-sensor contacts in $87.5\%$ of cases while reaching an average positional error in the order of $2$ centimeters. Our analysis also includes qualitative results of experiments with a real DIGIT sensor.
\end{abstract}

\section{INTRODUCTION}

The ability to estimate the 6D pose of objects is of paramount importance for autonomous robotic platforms that interact with the environment and the objects therein. A large number of methods from the computer vision and robotics communities have addressed the object pose estimation \cite{Xiang-RSS-18, tremblay2018deep, peng2019pvnet, song2020hybridpose, 9812299} and tracking \cite{wense3tracknet, 9363455, 9568706, stoiber2022iterative, 9811720} problems using RGB-D images of the scene as input.
\par
In robotic contexts, the sense of touch has also been employed to estimate or track the pose of objects, during in-hand manipulation, as the primary source of measurements \cite{sodhi2022patchgraph, 9562060, villalonga2021tactile, costanzo2021control, kelestemur2022tactile, sipos2022simultaneous,7378871} or complementary to vision \cite{7989460, 9709520, 9197117, dikhale2022visuotactile, 9561222} and sound \cite{gao2022objectfolder}. Research in this field is supported by the availability of new tactile sensing technologies that provide high resolution tactile images, including the most recent vision-based tactile sensors \cite{yuan2017gelsight, lambeta2020digit}, that we consider in this work.
\par
In the context of tactile-based object pose estimation, the majority of the work focuses on settings where the object is solely in contact with one sensor \cite{sodhi2022patchgraph, 7989460, gao2022objectfolder, 9709520, 9562060} or is manipulated by a sensorized parallel gripper \cite{kelestemur2022tactile, dikhale2022visuotactile, 9561222, sipos2022simultaneous}. Multiple vision-based sensors in contact are considered in \cite{villalonga2021tactile}, however their experiments are limited to objects with size comparable to that of the sensor, hence producing object pose-distinctive features. To the best of our knowledge, no other attempts have been made to incorporate measurements from a generic number of vision-based tactile sensors, especially when considering objects of standard size, as we do.
\begin{figure}
	\centering
	\includegraphics[scale=0.17]{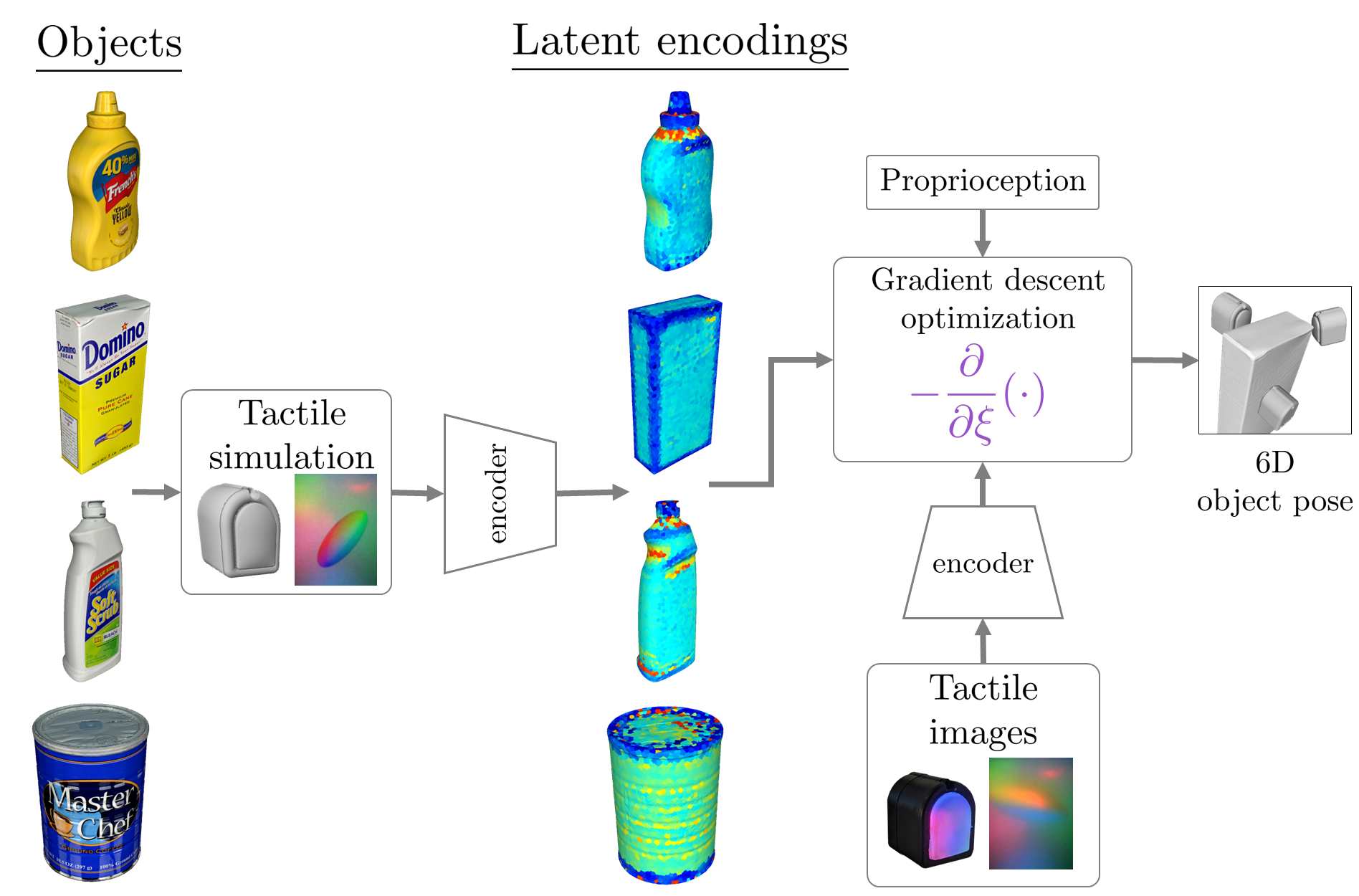}
	\caption{The proposed pipeline uses an object-agnostic CNN-based encoder to extract contact-related features on the surface of the object from simulated tactile images. We combine these features with tactile images and proprioception in order to estimate the in-hand 6D pose of an object in contact with multiple sensors.\label{fig:intro_figure}}
	\vspace{-2.0em}
\end{figure}
\par
In this work, we propose an optimization-based pipeline that incorporates proprioception and images from several vision-based tactile sensors in order to estimate the pose of an object in contact with them. We reason on the possible configurations of the sensors on the surface of the object 3D model and select the most appropriate according to geometric considerations and the compatibility with the images coming from the sensors. The compatibility is tested using a Convolutional Neural Network (CNN) trained on simulated tactile images. We then use gradient descent (GD) to find a suitable pose for the object by minimizing the distance between the real position of the sensors and the selected configurations.
\par
Differently from works dealing with small objects \cite{sodhi2022patchgraph, villalonga2021tactile}, i.e. having a size comparable to that of the sensor, we instead consider objects of standard size. One of our key insights is that, in such setting, the tactile images of the contact do not depend on the specific instance of the object, as it happens with small objects. Rather, the resulting contact patches can be approximated using simple shapes such as ellipsoidal and rectangular patches with varying aspect ratios.
We use this insight to train the CNN with object-agnostic data.
Our contributions are the following:
\begin{itemize}
    \item we propose a GD-based in-hand 6D object pose estimation method which combines several tactile images with proprioception;
    \item we devise an object-agnostic CNN-based encoding for the tactile images, and training procedure in simulation;
    \item we provide the results of simulated and real-world trials using objects from the standard YCB model set \cite{7251504Calli} and tactile images from a DIGIT tactile sensor\cite{lambeta2020digit};
    \item we compare against a purely geometric baseline showing the advantage of using the information encoded by the tactile images.
\end{itemize}

\section{RELATED WORK}
Our work is closely related to the recent literature on tactile-based and visuotactile-based in-hand 6D object pose estimation and tracking.
\par
\textbf{Visuotactile-based methods}
In \cite{7989460} the authors propose to extract point clouds from a single GelSight vision-based sensor and fuse them with point clouds from a depth camera using optimization based on signed-distance functions.
More recently, \cite{dikhale2022visuotactile} extends this idea by adding also RGB information and by fusing it with point clouds from depth and touch within a multi-stage CNN network. The CNN is trained with synthetic data obtained from the 3D meshes of several YCB \cite{7251504Calli} objects. 
A different path is considered in \cite{9197117} where a prior on the object pose from vision is refined using a GPU-accelerated contact physics simulator that is fed with contact data from a sensorized multifingered hand during in-hand manipulation.
\par
Unlike these works, this paper focuses on solely using tactile data without considering visual inputs.
\par
\textbf{Tactile-based methods}
In \cite{sipos2022simultaneous} the authors use Bayesian filtering to fuse proprioception and tactile feedback, in the form of force torque (FT) sensing, and estimate both the location of contacts and the in-hand object pose. 
\par
Other works focus specifically on vision-based tactile sensing. In \cite{sodhi2022patchgraph} an image translation network is trained, using real and simulated images, to reconstruct local object surface normals from tactile images. Normals are then used within a factor-graph to infer the object pose. Similarly, in \cite{villalonga2021tactile} the authors train a neural network with real object-agnostic data to reconstruct the local shape of the object at the contact point. The reconstructed images are then compared to a set of simulated images of the same object in order to find the most likely pose of the object. In \cite{kelestemur2022tactile} a CNN, trained on the ShapeNet dataset \cite{chang2015shapenet}, is integrated within a Bayesian filter in order to map the images from two vision-based tactile sensors, mounted on a gripper, to the position and orientation of the
gripper with respect to the object.
\par
Among these approaches, \cite{sodhi2022patchgraph, villalonga2021tactile} are different from our work as they focus on small objects \cite{sodhi2022patchgraph} or objects of comparable size to that of the sensor \cite{villalonga2021tactile}, resulting in \emph{local} tactile features that are distinctive of the object pose. Instead we consider objects of standard size, that produce \emph{local} features that are less dependant of the object pose. Moreover, in \cite{sodhi2022patchgraph} the object contacts one sensor only, while we consider multiple sensors.
\par
The authors of \cite{kelestemur2022tactile} make the same hypotheses on the size of objects as ours, however they limit the number of sensors to two, placed in a strict relative configuration, i.e. mounted on a gripper. Conversely, we do not make hypotheses on the sensors configuration. Moreover, they use meshes of real objects for training, while in our work we use simpler ellipsoidal and rectangular patches.
\par
Remarkably, \cite{7378871} is similar to our work as it uses a PCA-based descriptor to find object poses where the covariances of the local patches of the object 3D model match the principal components of the tactile data, represented as a 3D point cloud. In our work we use a similar concept but we substitute the PCA-based descriptor with CNN-based encoding that is suitable to process the output of vision-based tactile sensors in the form of RGB images. 

\section{Methodology}
The setting we consider consists of $L$ vision-based tactile sensors that are in contact with an object $\mathcal{O}$, with reference frame ${O}_{xyz}$. We assume that a 3D mesh model of the object and of the sensor are available.

Given  the $L$ sensors, their associated RGB images $\{I_{i}\}$ and poses $\{S^{w}_{i} \in \mathrm{SE}(3)\}$, in the world frame $W_{xyz}$, the proposed method estimates the pose of the object $T^{w} \in \mathrm{SE}(3)$. 
To this end, we instantiate a set of $N$ initial candidate poses $\{T_{j, 0}^{w} \in \mathrm{SE}(3)\}$ each associated with a $L$-tuple $t_{j}$

\begin{equation}\label{eq:tuple}
    t_{j} = \{ t_{1, j}^{o}, \cdots, t_{L, j}^{o}\},
\end{equation}
where $t_{i, j}^{o} \in \mathbb{R}^{3}$ is the \emph{candidate} position of the $i$-th sensor within the $j$-th candidate pose, expressed in the object reference frame $O_{xyz}$. These tuples are chosen according to the images $\{I_{i}\}$. Specifically, we use a CNN-based approach to favour positions $t_{i, j}^{o}$ that are compatible with the contact patch captured by the $i$-th image $I_{i}$.

The candidate poses are then iteratively refined via GD optimization in order to minimize the distance between the sensors positions $t_{i, j}^{w}(T_{j}^{w})$, expressed in the world frame according to the pose $T_{j}^{w}$, and the sensor poses $S_{i}^{w}$ .
Physical reasoning is finally leveraged in order to rank the resulting poses by giving more importance to those that are not in collision with the sensors. The best pose is chosen as the estimated pose $T^{w}$.

In the following we detail how we select the tuples $\{t_{j}\}$ given the images $\{I_{i}\}$ in Sec. \ref{sec:autoencoder}, we describe the GD-based optimization algorithm in Sec. \ref{sec:GD} and the physical reasoning-based ranking mechanism in Sec. \ref{sec:post-optim}. An overview of the proposed pipeline is depicted in Fig. \ref{fig:scheme}.

\subsection{Selection of the tuples}\label{sec:autoencoder}
We build a database of candidate sensor positions $t^{o} \in \mathbb{R}^{3}$ by sampling $M$ points uniformly in space along the surface of the 3D mesh of the object. For each point we collect a simulated tactile image $I(t^{o})$ by placing the sensor in position $t^{o}$ so that it is in contact with the object mesh with the normal to the object surface aligned with the normal out of the sensor surface. We use the Gazebo simulator \cite{Koenig-2004-394} and the DIGIT simulator TACTO \cite{wang2022tacto}
to collect the tactile image readings offline and we save them in a database, for each object of interest.
\par
\begin{figure*}
    \vspace{0.7em}
	\centering
	\includegraphics[scale=0.20]{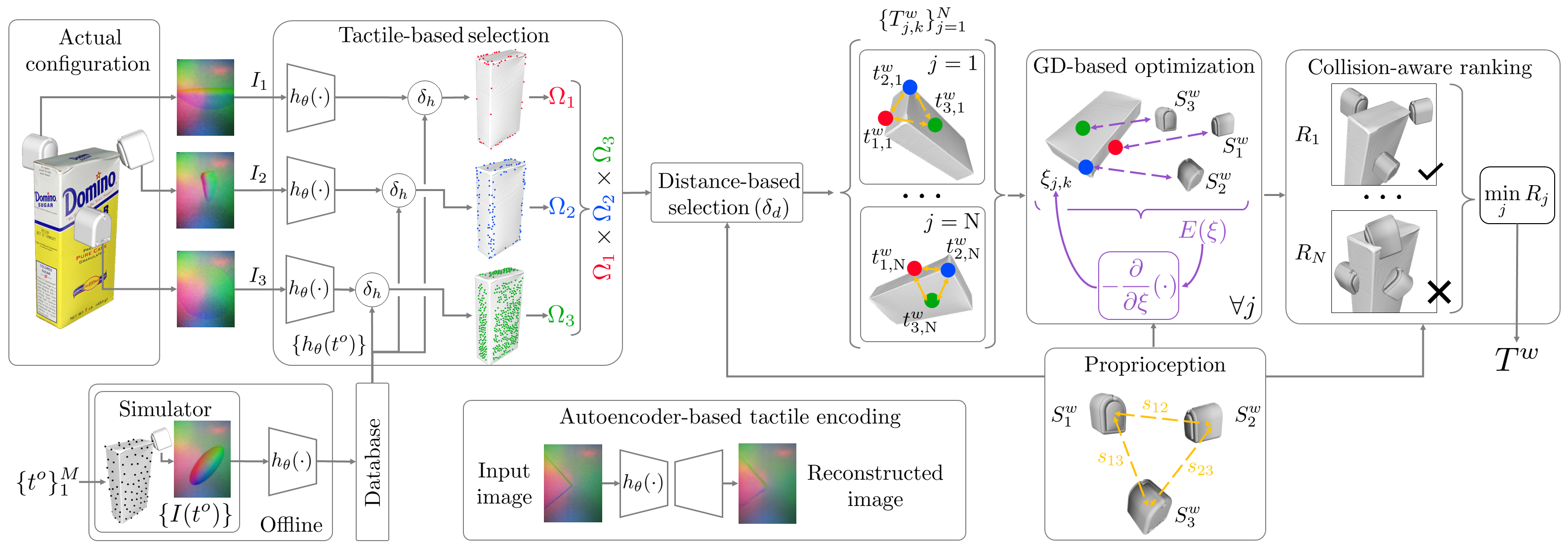}
	\caption{Overview of our pipeline for in-hand 6D object pose estimation using multiple vision-based tactile sensors. In the figure we consider the case in which 3 sensors are used, i.e. $L = 3$.\label{fig:scheme}}
	\vspace{-1.5em}
\end{figure*}
\textbf{Tactile-based selection}\label{sec:tactile_selection_mechanism}
The collected images are compared with the actual sensor images $\{I_{i}\}$ in order to identify which candidate sensor positions are compatible with the readings from the sensor. To this end, we use a latent encoding $h_{\theta}(I) \in \mathbb{R}^{128}$ corresponding to the encoder part of a CNN-based autoencoder, later detailed, that is trained to reconstruct tactile images. Here, $\theta$ indicates the weight of the autoencoder. Moreover, we define $I_{nc}$ as the simulated tactile image obtained when the sensor is not in contact with the object.
\par
We assume that a given sensor position $t^{o}$ is compatible with the image $I_{i}$ if
\begin{equation}\label{eq:tactile_based_criterion}
    |h_{\theta}(t^{o})^{T} h_{nc} - h_{\theta}(I_{i})^{T} h_{nc}| < \delta_{h},
\end{equation}
where $h_{\theta}(t^{o}) = h_{\theta}(I(t^{o}))$, $h_{nc} = h_{\theta}(I_{nc})$ are the features representing the absence of contact and $\delta_{h}$ is a threshold. I.e., we favour sensor positions whose tactile images have an inner-product distance (in latent space) from $h_{nc}$ similar to that of the image $I_{i}$. We collect all the compatible sensor positions for the $i$-th sensor in the set $\Omega_{i}$.
\par
In Fig. \ref{fig:intro_figure} we show an example of the latent encoding represented on top of the surface of several objects from the YCB \cite{7251504Calli} model set. In Fig. \ref{fig:scheme}, in the section ``Tactile-based selection'', we show an example of the sets $\Omega_{i}$ when the sensor is touching the ``Sugar box'' object within a corner or along an edge or a flat surface.
\par
Once the candidate sensor positions have been selected for each sensor $i$, we obtain the set of candidate tuples $\{t_{j}\}$, with $t_j$ as in Eq. \eqref{eq:tuple}, as the Cartesian product of the sets of all the sensors, i.e.:
\begin{equation}\label{eq:tuples}
    \{t_{j}\} = \Omega_{1} \times \cdots \times \Omega_{L}.
\end{equation}
\par
\textbf{Distance-based selection}\label{sec:dbs}
Before assigning each tuple in Eq. \eqref{eq:tuples} to a candidate pose $T_{j}^{w, 0}$, we exploit the knowledge of the sensors poses $S_{i}^{w}$ to filter out all the tuples whose sensor positions do not respect the relative distance between the actual sensor poses $S_{i}^{w}$.
\par
We define $s_{i} \in \mathbb{R}^{3}$ as the translational part of the homogeneous transform $S_{i}^{w}$. Hence, we keep a tuple $t_{j}$ if the following holds:
\begin{equation}\label{eq:distance_based_criterion}
    \frac{2}{L (L-1)}\sum_{i=1}^{L-1} \sum_{k=i+1}^{L} \left|\|s_{i} - s_{k}\| - \|t_{i, j}^{o} - t_{k, j}^{o}\|\right| < \delta_{d}.
\end{equation}
In order to keep only a predetermined number of candidate poses $N$, we iteratively reduce $\delta_{d}$ until the number of tuples is less or equal to $N$.
\begin{figure}
	\centering
	\includegraphics[scale=0.23]{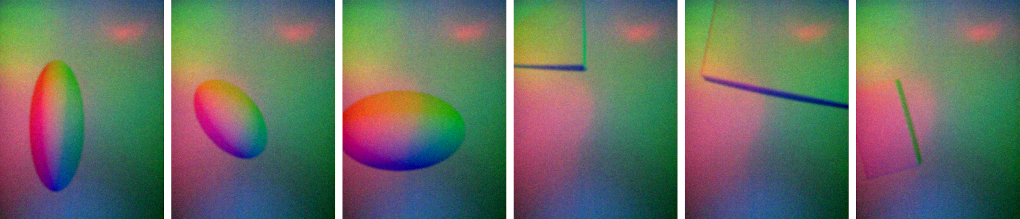}
	\caption{Sample training images used to train the autoencoder for tactile image reconstruction.\label{fig:training}}
	\vspace{-1.5em}
\end{figure}

\textbf{Autoencoder training and details}
In our setting we assume that the objects of interest occupy a much larger volume than that of the sensor, as it is usual when comparing a fingertip with an object of standard size. As a consequence, we found that the resulting tactile patches do not depend on the specific object instance and are adequately represented by ellipsoidal and rectangular shapes of diverse aspect ratios. Therefore, we train the autoencoder to reconstruct object-agnostic simulated images of this kind, without the necessity to collect training images along the surface of each object of interest. 
\par
In practice, we obtained the images by simulating the contact with 3D models of superquadrics \cite{barr1981superquadrics} with varying parameters, so as to obtain diverse shapes from spheres to ellipsoids to prisms and different aspect ratios. Fig. \ref{fig:training} shows several examples of training images.
\par
Our network is inspired by the autoencoder for implicit 3D object rotation encoding presented in \cite{Sundermeyer_2018_ECCV}. It comprises 4 convolutional and 4 deconvolutional layers for the encoder and decoder networks, respectively, with ReLU activation functions. We train the network by minimizing the pixel-wise L2 distance between the input image and the decoder output.

\begin{table*}
\vspace{0.7em}
\tiny
\scriptsize
\centering
\caption{Results of the experiments in simulation when using three sensors: positional, ADI-AUC errors and contact accuracy for several methods. \#1 and \#1-5 indicate results evaluated on the best pose and averaged on the best five poses, respectively.\label{tab:exps_sim_2000_3}}
\begin{tabular}{| l | c | c | c | c | c | c | c | c | c | c|}
\hline
Metric & \multicolumn {2}{c|}{Positional error (cm) $\downarrow$} & \multicolumn {2}{c|}{ADI-AUC$_{2 \mathrm{cm}}$ (\%) $\uparrow$} & \multicolumn {2}{c|}{ADI-AUC$_{2 \mathrm{cm}}$\,(rotation)\,(\%) $\uparrow$} & \multicolumn{2}{c|}{Contact accuracy (\%) $\uparrow$}\\
\hline
Method & Baseline & Ours & Baseline & Ours & Baseline & Ours & Baseline & Ours\\
\hline
Evaluated poses & \#1, \#1-5 & \#1, \#1-5 & \#1, \#1-5 & \#1, \#1-5 & \#1, \#1-5 & \#1, \#1-5 & \#1, \#1-5 & \#1, \#1-5\\
\hline
002{\_}master{\_}chef{\_}can & 5.02, 4.96 & \bb{1.27}, \bb{1.26} & 25.00, 23.78 & \bb{72.97}, \bb{84.23} & 96.31, 93.19 & \bb{96.36}, \bb{94.79} & 25.00, 10.00 & \bb{100.00}, \bb{100.00}\\
004{\_}sugar{\_}box & 4.01, 3.60 & \bb{2.05}, \bb{1.87} & 25.00, 35.42 & \bb{70.69}, \bb{66.95} & 69.78, 61.50 & \bb{71.81}, \bb{80.28} & 0.00, 20.00 & \bb{100.00}, \bb{100.00}\\
006{\_}mustard{\_}bottle & 3.15, 3.09 & \bb{1.58}, \bb{1.50} & 46.06, 48.10 & \bb{71.62}, \bb{77.99} & 88.90, 77.33 & \bb{93.69}, \bb{86.98} & 25.00, 35.00 & \bb{100.00}, \bb{100.00}\\
007{\_}tuna{\_}fish{\_}can & \bb{1.48}, \bb{1.47} & 2.17, 1.84 & 94.94, 92.12 & \bb{95.05}, \bb{93.52} & 95.52, 93.96 & \bb{97.44}, \bb{95.45} & 50.00, 35.00 & \bb{75.00}, \bb{65.00}\\
008{\_}pudding{\_}box & 2.78, 2.47 & \bb{1.97},\bb{2.10} & \bb{92.05}, \bb{88.06} & 71.28, 73.27 & 95.21, \bb{92.86} & \bb{95.92}, 88.11 & 25.00, 20.00 & \bb{100.00}, \bb{90.00}\\
011{\_}banana & 4.07, \bb{3.46} & \bb{3.23},3.65 & 46.58, \bb{43.96} & \bb{66.48}, 42.47 & \bb{71.86}, \bb{59.08} & 69.47, 58.46 & 25.00, 30.00 & \bb{75.00}, \bb{55.00}\\
019{\_}pitcher{\_}base & 9.69, 9.54 & \textbf{4.81}, \textbf{4.63} & 0.00, 0.00 & \textbf{25.00}, \textbf{35.14} & 0.00, 0.00 & \bb{25.00}, \bb{35.09} & 0.00, 5.00 & \bb{100.00}, \bb{95.00}\\
021{\_}bleach{\_}cleanser & 5.23, 5.65 & \bb{2.20}, \bb{2.63} & 0.00, 34.72 & \bb{91.31}, \bb{66.64} & 46.66, 67.33 & \bb{94.29}, \bb{91.05} & 25.00, 30.00 & \bb{75.00}, \bb{95.00}\\
036{\_}wood{\_}block & 9.95, 10.41 & \bb{3.27}, \bb{3.49} & 0.00, 5.00 & \bb{47.09}, \bb{30.34} & 67.12, \bb{61.95} & \bb{70.59}, 57.58 & 25.00, 20.00 & \bb{75.00}, \bb{80.00}\\
040{\_}large{\_}marker & 2.26, 2.19 & \bb{1.05}, \bb{0.91} & 73.01, 75.48 & \bb{96.89}, \bb{77.10} & 74.01, 78.32 & \bb{97.89}, \bb{83.01} & 50.00, 50.00 & \bb{75.00}, \bb{90.00}\\
\hline
Mean & 4.76, 4.68 & \bb{2.36}, \bb{2.39} & 40.26, 44.66 & \bb{70.84}, \bb{64.77} & 70.54, 68.55 & \bb{81.25}, \bb{77.08} & 25.00, 25.50 & \bb{87.50}, \bb{87.00}\\
\hline
\end{tabular}
\end{table*}

\subsection{GD-based 6D object pose optimization}\label{sec:GD}
Given an initial candidate pose $T_{j, 0}^{w}$, and the associated tuple $t_{j}$,
we iteratively refine the object pose $T_{j,k}^{w}$ using a gradient descent-based approach:
\begin{equation}
    \xi_{j}^{k+1} = \xi_{j}^{k} -K \frac{\partial E(\xi)}{\partial \xi}\biggr\rvert_{\xi_j^k},
\end{equation}
with $\xi_{j}^{k} \in \mathbb{R}^{6}$ such that 
\begin{equation}
    T^{w,k}_{j} = T^{w,k}_{j}(\xi_{j}^{k}) = 
    \begin{bmatrix}
    \mathrm{exp}((\xi_{j\,(3:6)}^{k})^{\wedge}) & \xi_{j\,(1:3)}^{k}\\
    0^{T} & 1\\
    \end{bmatrix},
\end{equation}
where $(\xi_{j\,(3:6)}^{k})^{\wedge} \in \mathfrak{s}\mathfrak{o}(3)$ is an element in the Lie-algebra of $\mathrm{SO}(3)$. Moreover, $K \in \mathbb{R}^{6 \times 6}$ is a tuning matrix and $E(\xi)$ is the loss
\par
\begin{equation}\label{eq:loss}
    E(\xi) = \frac{1}{L}\sum_{i = 1}^{L} \|s_{i}(S_{i}^{w}) - t^{w}_{i}(T(\xi))\|^{2},
\end{equation}
where $t_{i}^{w}$ is the $i$-th sensor candidate position at the pose $T(\xi)$, expressed in the world frame. In summary, the loss in Eq. \eqref{eq:loss} is used to minimize the distance between the candidate sensor positions and the actual ones.
\par
In order to initialize the iterates, we set the rotational part of $T_{j,0}^{w}$ to a matrix uniformly sampled on the group of the rotations $\mathrm{SO}(3)$. As regards to the translational part, in practice we consider not just one hypothesis but several. Specifically, we sample 14 points at distance $l_s$ from the geometric mean of the sensors positions $s_c = (s_1 + \cdots + s_L)/L \in \mathbb{R}^{3} $ in the directions of the vertexes and face centers of a cube centered in $s_c$. 
We also include the center $s_c$ as one of the hypotheses. This choice is made to better explore the state space at initialization time.
\par
We run the GD optimization routine in parallel for all the $N$ poses for a maximum of $k_{max}$ steps.

\subsection{Physical reasoning-based pose ranking}\label{sec:post-optim}
Once the optimization process is done, we rank the resulting poses $\{T_{j, k_{max}}^{w}\}_{j=1}^{N}$ according to their loss and by penalizing poses where the object mesh is in collision with the mesh of the sensors, as they correspond to unfeasible configurations.
\par
To this end, we use a physics engine in order to evaluate the penetration depths $d_{i, j}$ between the object mesh for each pose $T_{j, k_{max}}^{w}$ and the mesh of the sensor at each pose $S_{i}^{w}$. We then define the ranking score as follows:
\begin{equation}\label{eq:ranking}
    R_j = E(\xi_{j}^{k_{max}}) + \underset{i}{\max}\,d_{i,j},
\end{equation}
i.e. we sum the GD final loss with the highest penetration depth between the object and the sensor meshes. We finally select the object pose $T^{w}$ as the one with the least ranking score:
\begin{equation}
    \begin{split}
        T^{w} &= T^{w}_{j^{*}, k_{max}},\\
        j^{*} &= \underset{j}{\argmin}\,R_{j}.
    \end{split}
\end{equation}

\section{Experimental results}

In this section we present the results of several experiments carried out both in simulation and in a real setting using the DIGIT vision-based tactile sensor \cite{lambeta2020digit}. We compare both quantitatively and qualitatively against a purely geometric baseline that is obtained by running our algorithm without the tactile-based selection mechanism described in Sec. \ref{sec:tactile_selection_mechanism}. In this way, the
baseline acts as a point-set registration algorithm that tries to align the candidate sensors positions with the actual sensors positions by using only proprioception. Although our work is similar to \cite{kelestemur2022tactile}, we could not compare to it as the software implementation is not available.
\par
We use PyTorch \cite{NEURIPS2019_9015} to train the autoencoder for tactile image reconstruction, the JAX \cite{jax2018github} Python library to perform the GD-based optimization in parallel and the NVidia PhysX \cite{physx} middleware to evaluate the collision depths between object and sensors. Our software implementation is made publicly available for free with an Open Source license online\footnote{https://github.com/hsp-iit/multi-tactile-6d-estimation}.
\par
For all the experiments, we used the following values for the parameters of the algorithm: $N = 5000$, $l_{s} = 0.2\,\mathrm{m}$, $k_{max} = 700$, $M = 2000$, $K = \mathrm{diag}(I_{3} 10^{-2}, I_{3})$.

\begin{figure*}
    \vspace{0.7em}
	\centering
	\includegraphics[scale=0.22]{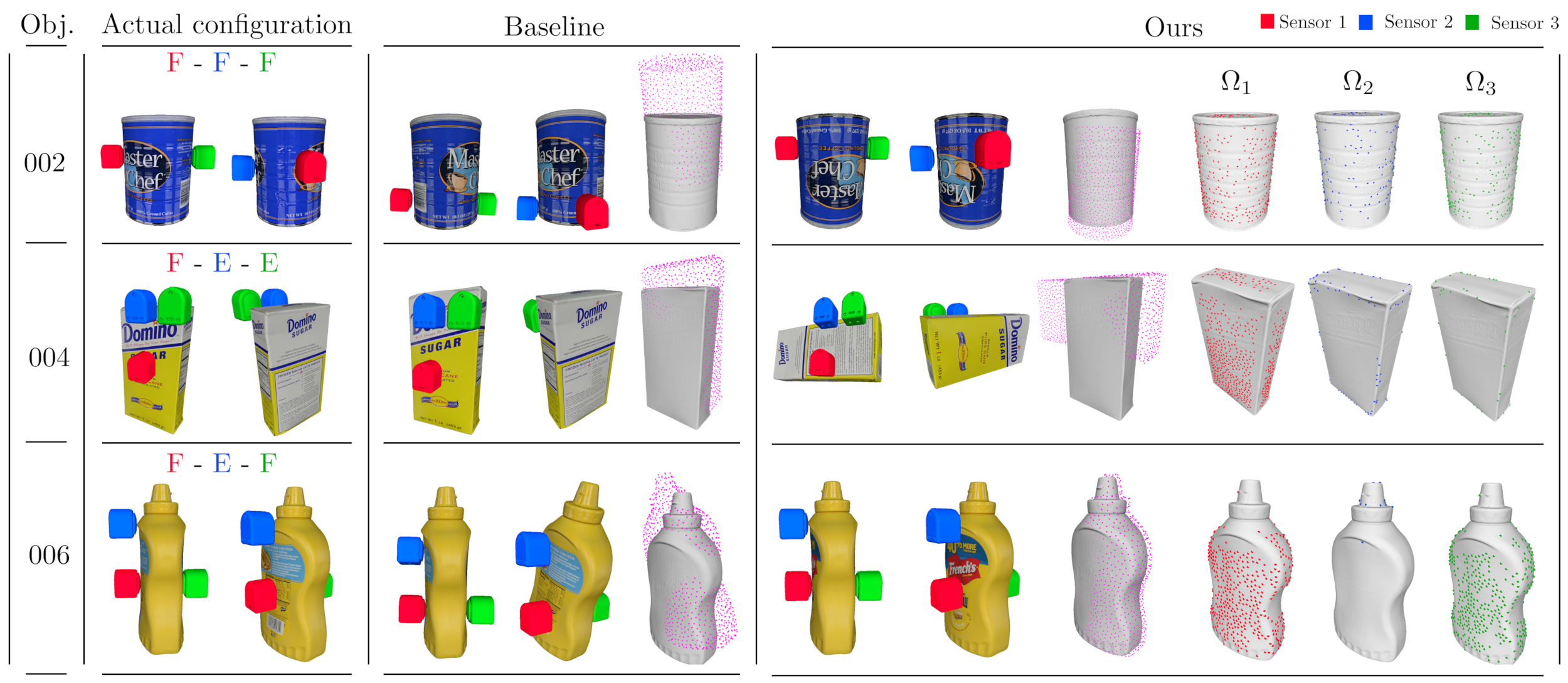}
	\caption{Qualitative results in simulation for the objects ``002\_master\_chef\_can'', ``004\_sugar\_box'' and ``006\_mustard\_bottle''. The letter ``F'' indicates contact with a flat surface, while ``E'' indicates contact along an edge of the object In this scenario, we use $L = 3$ tactile sensors.\label{fig:qual_sim}}
	\vspace{-1.8em}
\end{figure*}

\subsection{Experimental setup}
\textbf{Simulated setup} We consider 10 objects from the YCB model set \cite{7251504Calli}. For each object, we select several object-sensor configurations with three sensors. The configurations are selected in order to explore several kind of contacts, e.g. with flat surfaces, edges, corners and rounded parts of object, when available. We remark that the poses of the sensors in the above configurations are different from those used to collect the database described in Sec. \ref{sec:autoencoder}. For each configuration, we place the object of interest within the Gazebo simulator at a known fixed pose, used as ground truth $T^{w, gt}$, and we collect the $L$ sensors poses $S_{i}^{w}$ and images $I_{i}$. Sample configurations are shown in Fig. \ref{fig:qual_sim}. \par We evaluate the performance of our method against the baseline by considering the error in position with respect to the ground truth. We do not provide a pure rotational error as it would not be suitable in the presence of objects with symmetries.
Instead, we include the standard ADI-AUC metric \cite{Xiang-RSS-18}, with threshold set at $2\,\mathrm{cm}$, that jointly evaluates the positional and rotational errors while taking into account the symmetries of the objects. We also provide the ADI-AUC metric evaluated by setting the estimated position equal to the ground truth, in order to report on the rotational error separately. Moreover, we evaluate the ability of our method to estimate object poses that are compatible with the actual contacts. To this end, we report on the percentage of experiments where the object, at the estimated pose, is in contact with all the sensors in parts of the surface that are compatible with the real object-sensors configuration. We remark that, for each object, the results are averaged on all the experiments involving that object.
\par
\begin{figure}
	\centering
	\includegraphics[scale=0.9]{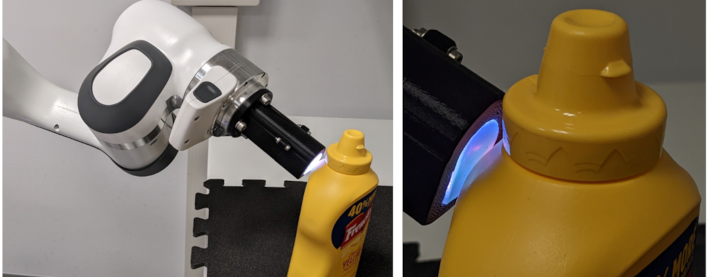}
	\caption{Real world experimental setup. On the left: the Franka Emika Panda robot with a DIGIT sensor mounted at the end-effector. On the right: a detail of the DIGIT sensor in contact with the object.\label{fig:real_setup}}
	\vspace{-2.0em}
\end{figure}

\textbf{Real world setup} For the experiments in a real scenario, we use a DIGIT sensor mounted on a 7-DoF Franka Emika Panda robot (see Fig. \ref{fig:real_setup}). We use the robot to touch the object with the sensor several times in order to reproduce the scenario of simultaneous contacts. The sensor 6D poses are obtained using the robot forward kinematics.
In this case we used the Taxim example-based simulator \cite{9681378} to simulate the tactile images required for training and to build the database. In our experiments, Taxim provided latent features that are better suited for this scenario, thanks to the calibration procedure that make rendered images more similar to those of the real sensor.

\subsection{Results of the experiments in simulation}

\textbf{Pose accuracy}
Table \ref{tab:exps_sim_2000_3} reports the positional error and the ADI-AUC metrics for the best pose and the five best poses, in the case $L = 3$. In the following we discuss the results taking into account the best pose.
\par
Considering all the objects on average, our method achieves the best performance according to all the metrics and gets an average positional error in the order of few centimeters. The positional error for the baseline is almost doubled on average. Moreover, the maximum error for the baseline almost reaches $10\,\mathrm{cm}$, for the objects ``021'' and ``036'', while with our method it stays below $5 \,\mathrm{cm}$.
\par
According to the ADI-AUC and the ADI-AUC (rotation) metrics, our method provides an increase in performance of approximately $76 \%$ and $15 \%$ on average, respectively.
\par
\textbf{Contact accuracy}
Table \ref{tab:exps_sim_2000_3} reports the contact accuracy for both methods. As can be seen, our method outperforms the baseline on average and for each object separately. On average, our method increases the performance by $250 \%$. This suggests that, even though the baseline might achieve a reasonable pose accuracy, most of the times the estimated poses are not compatible with the actual contacts between the object and the sensors. In this respect, we can consider our method as a way to boost the performance of a point-set registration-like algorithm using tactile information.

\par
\textbf{Qualitative results} In Fig. \ref{fig:qual_sim}, we provide qualitative results for several YCB objects. For each row, we provide two views of the actual object-sensors configuration and similar ones showing the estimate provided by the baseline and our method. We also show a direct comparison between the ground truth pose, represented with the solid gray mesh without texture, and the estimated pose, represented with the magenta point cloud. In the last three columns, the sets of candidate contact points $\Omega_{i}$ are reported for each sensor.
\par
\begin{figure}
    \vspace{0.7em}
	\centering
	\includegraphics[scale=0.43]{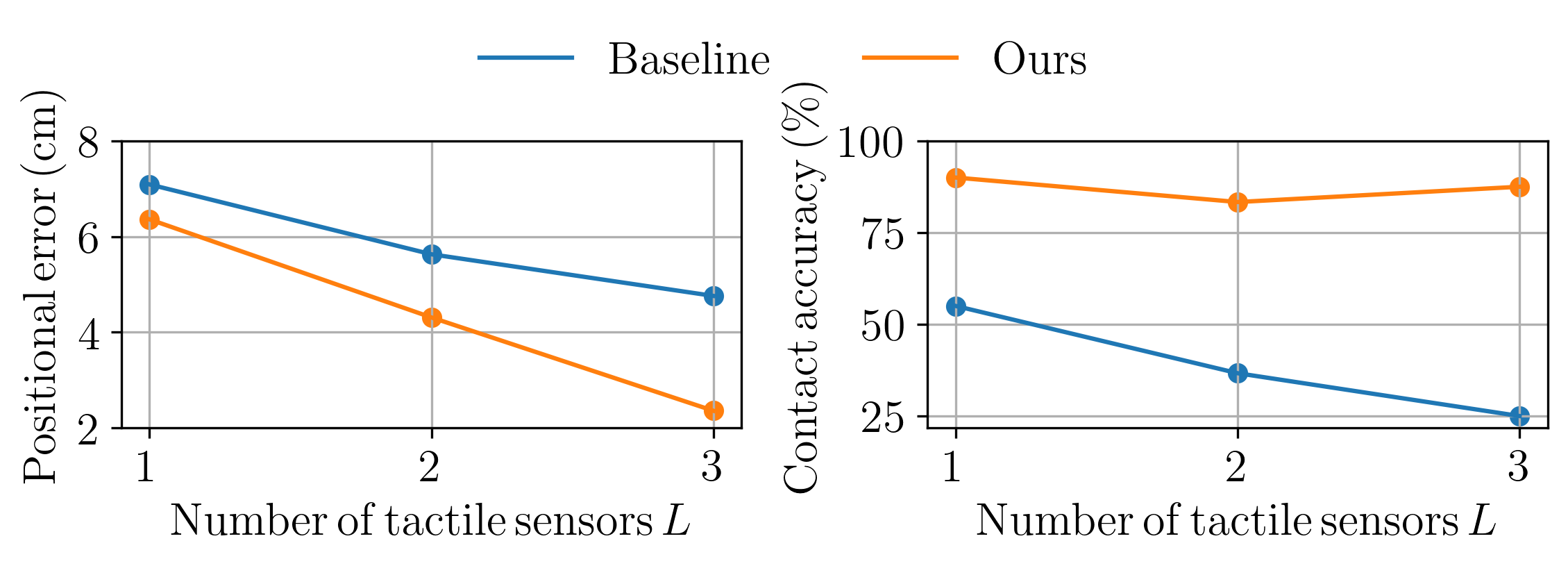}
	\caption{Positional errors and contact accuracy when the number of tactile sensors varies from one to three.\label{fig:number_contacts}}
	\vspace{-2.0em}
\end{figure}
Although the baseline can achieve reasonable pose estimates, in all the presented examples at least one sensor contacts the object in a part of the surface that is not compatible with actual object-sensors configuration. E.g., for the object ``002'', the second sensor should touch a flat surface, while it is in contact with an edge. For the object ``004'', both the second and third sensors should touch an edge, while they are in contact with a flat surface. A similar reasoning holds for the object ``006''. 
\par
On the other hand, the pose estimates provided by our algorithm are always compatible with the actual contacts. We remark that, for the object ``004'', the estimated rotation is quite different from the ground truth. This depends on the multimodality of the object poses given the measurements at disposal, i.e. the pose of the sensors and the tactile images.

\par
\textbf{Role of the number of sensors $\mathbf{L}$} In Fig. \ref{fig:number_contacts}, we report on the performance of the algorithms when the number of tactile sensors varies from one to three. For both methods, the positional error decreases as the number of sensors increases. As regards the contact accuracy, while it is almost constant for our method, it decreases for the baseline when more sensors are used. This fact can be explained by the increase in the number of possible configurations of the sensors. While our method is able to filter them according to the tactile images, the baseline considers all of them, thus reducing the probability to find those that are compatible with the actual contacts between the object and the sensors.

\subsection{Results of the experiments on the real setup}
\begin{figure}
   \vspace{0.7em}
	\centering
	\includegraphics[scale=0.16]{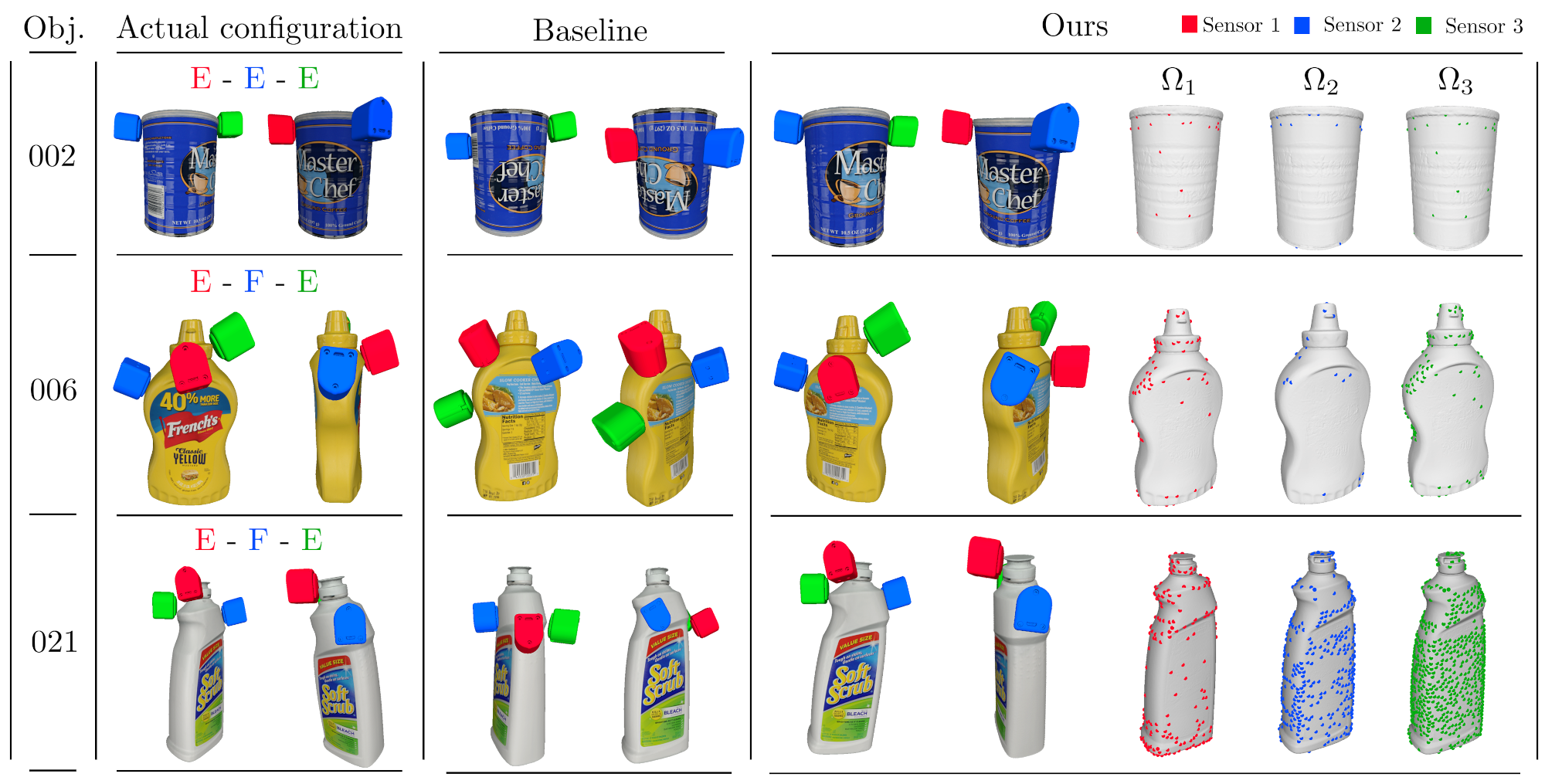}
	\caption{Qualitative results on the experiments from the real setup for the objects ``002\_master\_chef\_can'', ``006\_mustard\_bottle'', and
	``021\_bleach\_cleanser''.
 \label{fig:qual_real}}
	\vspace{-2.0em}
\end{figure}

In Fig. \ref{fig:qual_real}, we show qualitative results of the experiments from the real setup. As in Fig. \ref{fig:qual_sim}, we show several views of the object at the estimated pose alongside the sensors, whose pose is provided by the robot forward kinematics. Instead, we do not compare against the ground truth object pose, as not available in this scenario.
\par
When comparing against the baseline, our method estimates contact positions that are more compatible to the actual case. For example, in the case of object ``006'', the three sensors are touching an edge, a flat surface and another edge, respectively. In the case of the baseline, on the contrary, the three sensors are estimated to be in contact with a flat surface, an edge and another flat surface.

\subsection{Computation times}
The mean time required to execute the full pipeline of our method, averaged on all the experiments that we considered, corresponds to $10.2$s, $29.8$s and $35.7$s when using one, two or three sensors, respectively. For the baseline, these times increase to $14.5$s, $43.6$s and $71.9$s. This outcome is expected as, for the baseline, the distance-based selection criterion in Eq. \eqref{eq:distance_based_criterion} needs to be checked for all the possible combinations of candidate sensor positions, i.e. $M^{L}$. 

\subsection{Limitations}
At the moment, we do not completely exploit the information given by the sensor image, such as the portion of the sensor in contact, which could help mitigate the multimodality of the object poses given the measurements.
\par
Secondly, the selection mechanism of the sets $\Omega_{i}$ can result in overly sparse or dense sets depending on the choice of the parameter $\delta_{h}$, which needs an empirical tuning procedure.
\par
In the real experiments, the sets $\Omega_{i}$ might differ from the expected ones due to differences between real and simulated images. For instance, the set $\Omega_{2}$ for the object ``006'' in Fig. \ref{fig:qual_real}, corresponding to a flat surface, is much more sparse than the ideal one.
Conversely, numerous outliers can be found while touching an edge, as in the set $\Omega_{3}$ of object ``021".
\par
We finally remark that the proposed pose ranking criterion (see Sec. \ref{sec:post-optim}) is based on the Cartesian distance between measured and optimized sensor positions and on the penetration depth. As a consequence, it does not necessarily reward poses corresponding to a lower 6D pose error (as for the object ``004" in Fig. \ref{fig:qual_sim}). This can also be seen in Table \ref{tab:exps_sim_2000_3}, where the results of the best pose (\#1) are not always better than the averaged ones (\#1-5), as one might expect.

\section{Conclusion}
In this work, we presented an in-hand 6D object pose estimation method which combines proprioception with a generic number of vision-based tactile sensors. The core of the method relies on a latent encoding provided by an autoencoder that is trained completely in simulation. Experiments in simulation and in a real scenario demonstrate that the system allows us to estimate the pose of an object with touch alone, improving performance with respect to a geometric baseline. 
\par
As future work, we plan to better exploit the sensor information to reduce the multimodality and the estimation error even further. Moreover, we aim to use the pipeline for tracking purposes by using a stream of sensor images while the robot manipulates the object.

\addtolength{\textheight}{-1.0cm}  

\bibliography{biblio}

\end{document}